# The Trilingual Triad Framework: Integrating Design, AI, and Domain Knowledge in No-code AI Smart City Course

**Qian Huang**
Research Fellow, Lee Kuan Yew Center for Innovative Cities, Singapore University of Technology and Design

**King Wang Poon**
Director, Lee Kuan Yew Center for Innovative Cities; Chief Strategy and Design AI Officer, Singapore University of Technology and Design

## Abstract

This paper introduces the "Trilingual Triad" framework, a model that explains how students learn to design with generative artificial intelligence (AI) through the integration of Design, AI, and Domain Knowledge. As generative AI rapidly enters higher education, students often engage with these systems as passive users of generated outputs rather than active creators of AI-enabled knowledge tools. This study investigates how students can transition from using AI as a tool to designing AI as a collaborative teammate. The research examines a graduate course, Creating the Frontier of No-code Smart Cities at the Singapore University of Technology and Design (SUTD), in which students developed domain-specific custom GPT systems without coding. Using a qualitative multi-case study approach, three projects - the Interview Companion GPT, the Urban Observer GPT, and Buddy Buddy - were analyzed across three dimensions: design, AI architecture, and domain expertise. The findings show that effective human-AI collaboration emerges when these three "languages" are orchestrated together: domain knowledge structures the AI's logic, design mediates human-AI interaction, and AI extends learners' cognitive capacity. The Trilingual Triad framework highlights how building AI systems can serve as a constructionist learning process that strengthens AI literacy, metacognition, and learner agency.

## Introduction

The rapid proliferation of generative artificial intelligence has prompted a fundamental re-evaluation of its role in higher education. While early interventions focused on teaching students to use AI tools effectively, a new frontier is emerging that centers on empowering them to create and customize these tools themselves. This shift represents a move from a user-centric paradigm to one of creation, positioning students not merely as consumers of technology but as its architects. At the Singapore University of Technology and Design (SUTD), this pedagogical evolution is exemplified in the 'Creating the Frontier of No-code

Smart City' course, where students undertake the challenge of designing custom GPT-based knowledge tools tailored to specific learning objectives. This endeavor marks a significant departure from traditional educational practices where technology serves as a peripheral resource or a standardized application.

The central research goal of this study is to analyze the transformation of students from passive AI users to active creators of custom GPT-based knowledge tools within this course. The investigation focuses on understanding the mechanisms that drive this change, specifically the interplay between Design, Artificial Intelligence (AI), and Domain Knowledge across three detailed case studies. Each case study—a multimodal Interview Companion GPT, a structured Urban Observer GPT, and an interactive Buddy Buddy GPT— is examined through both deep-dive analyses and a comprehensive cross-case synthesis. This dual approach allows for a granular examination of the unique characteristics of each project while also uncovering overarching principles that govern the learning process. The transformation is framed by the "Tool-to-Teammate" instructional strategy, a pedagogical framework designed to guide students through this complex journey of co-design and implementation.

A critical component of this investigation is its grounding in a multi-theoretical framework. While Self-Determination Theory (SDT) offers a powerful lens for understanding the motivational dynamics at play, it is complemented by Constructionism, which explains the power of learning through making; the TPACK framework, which maps the integration of technological, pedagogical, and content knowledge; and Distributed Cognition, which conceptualizes the human-AI system as a unified cognitive entity. Together, these theories provide a comprehensive explanation for why the act of creating AI teammates is such a potent driver of deep learning and agency. This research explores how this multi-theoretical foundation manifests in practice, fostering a deeper, more sustainable form of engagement.

The significance of this study lies in its potential to provide a replicable and theoretically grounded model for integrating no-code AI development into higher education curricula. As generative AI becomes increasingly embedded in higher education and professional workflows, Scholars have emphasized the need to move beyond surface-level tool use toward pedagogical models that foster critical engagement, transparency, and learner agency (Kasneci et al., 2023; Zawacki-Richter et al., 2019; Willems et al., 2025). The 'Creating the Frontier of No-code Smart City' course at SUTD offers a compelling case study of how this can be achieved. By focusing on the synergistic relationship between domain expertise, AI architecture, and thoughtful design, this research illuminates a pathway toward fostering what can be termed "AI literate designers"—individuals who possess not only the cognitive knowledge and practical skills to work with AI but also the metacognitive awareness to do so ethically and effectively. The intended audience for this research includes academic scholars in the fields of learning sciences, human-computer interaction, and AI in education, offering them evidence-based insights into the design of future learning environments that leverage AI as a medium for student empowerment and deep, disciplinary learning.

**Literature Review**

The integration of artificial intelligence into education is a rapidly evolving field characterized by both significant opportunities and profound challenges. Research in this area spans two primary domains: "learning with AI," which involves using AI to augment teaching and learning processes, and "learning about AI," which focuses on educating students about the principles, applications, and societal implications of AI itself. This study falls squarely within the latter, exploring how students learn about AI by actively creating it. The literature on AI in education highlights a growing consensus that simply teaching students to use AI tools is insufficient for preparing them for an AI-shaped world. There is a pressing need to develop frameworks that promote deeper forms of engagement and critical thinking.

A key concept emerging from recent scholarship is AI literacy, which has been conceptualized as a multi-faceted competency encompassing awareness of AI systems, the ability to use and adapt them for specific tasks, critical evaluation of their outputs, and ethical understanding of their implications (Long & Magerko, 2020; Ng et al., 2021).One comprehensive framework defines AI literacy through four key components: Awareness (the ability to identify and comprehend AI technologies), Usage (proficiency in implementing AI for specific tasks), Evaluation (the capacity to critically analyze AI applications and their outcomes), and Ethics (an understanding of the responsibilities and risks associated with AI). Developing such literacy requires moving beyond simple operational skills like prompt engineering to include critical thinking and ethical reflection. The pedagogical approach described in this paper aligns directly with this vision, as the process of building custom GPTs necessitates that students engage with all four components of AI literacy simultaneously.

The theoretical underpinnings of this pedagogical approach are drawn heavily from Self-Determination Theory (SDT). Self-Determination Theory (SDT) emphasizes that learners' autonomy, competence, and relatedness are foundational psychological needs that strongly influence motivation, engagement, and persistence in technology-enhanced learning environments, including AI-supported education (Holmes et al., 2019; Zawacki-Richter et al., 2019). Recent scholarship has applied SDT extensively to explain motivation in technology-enhanced learning environments, particularly concerning AI. Studies have shown that when teachers support these basic needs, students experience enhanced competence and autonomy, which positively impacts their psychological health and learning processes. Furthermore, satisfying these psychological needs has been found to be a pivotal factor in advancing students' AI literacy, suggesting a direct link between motivational climate and technical competency. This body of research provides a strong foundation for analyzing how the "Tool-to-Teammate" strategy in this course is designed to foster student agency by systematically supporting these fundamental needs.

The practice of having students create their own AI tools aligns with the broader movement towards participatory and human-centered design in education. Emerging studies suggest that students deepen AI literacy when they move from using generative AI to designing customized AI systems tailored to domain-specific tasks (Huang & Willems, 2025; Huang et al., 2025a).Participatory design emphasizes an inclusive, iterative process where stakeholders, including learners, are active co-designers rather than passive recipients of a finished product. This approach values shared understanding and distributed decision-making, fostering democratic engagement. Similarly, Human-Centered AI (HCAI) emphasizes the design of AI systems that augment human capabilities, preserve human agency, and ensure meaningful human control, particularly in high-stakes domains such as education (Shneiderman, 2020; Kasneci et al., 2023). Empirical studies have also begun to document how human–AI collaboration can enhance professional practice in applied learning domains such as fitness education and design education (Huang & Poon, 2025; Huang et al., 2025b).HCAI in education aims to create personalized and equitable learning experiences, positioning AI as a partner in the learning process. The student-led development of GPTs in the SUTD course embodies these principles, as students are not just using a pre-designed tool but are actively shaping its functionality, interface, and underlying logic to meet their specific pedagogical needs. This process transforms students from mere tool-users into designers and collaborators, fundamentally altering their relationship with the technology they are learning about. This evolution from AI as a tool to AI as a teammate is supported by emerging evidence that well-designed human–AI partnerships can enhance learners' self-efficacy, agency, and engagement, particularly when roles between humans and AI are deliberately complementary (Holstein et al., 2019; Celik et al., 2023).

**Theoretical Framework**

The pedagogical framework guiding the course is built upon a robust, multi-theoretical foundation that explains the cognitive, motivational, and social dimensions of the "Tool-to-Teammate" strategy. This framework integrates four complementary theories: Self-Determination Theory (SDT), Constructionism, the Technological Pedagogical Content Knowledge (TPACK) framework, and Distributed Cognition.

Self-Determination Theory (SDT) provides the motivational core of the framework. SDT posits that intrinsic motivation and optimal learning are fostered when individuals' fundamental psychological needs for autonomy, competence, and relatedness are satisfied. In the context of this course, the act of creating a custom GPT directly addresses these needs. Autonomy is realized as students gain control over their learning technologies; competence is built through the mastery of complex design and technical challenges; and relatedness is fostered through collaborative design processes. This motivational engine drives deep engagement and a sense of ownership over the learning process.

Constructionism, as proposed by Seymour Papert, complements SDT by explaining the cognitive power of the "making" process itself. Contemporary learning sciences research has extended constructionist principles to digital and AI-mediated environments, arguing that learning is deepened when learners actively construct meaningful artifacts that externalize and formalize their understanding (Bialik & Fadel, 2018; Holmes et al., 2019). The student-built GPTs are precisely these kinds of constructionist artifacts. The process of designing and building a functional AI teammate forces students to grapple with and synthesize complex ideas from multiple domains. Recent scholarship argues that generative AI naturally aligns with constructionist principles by enabling personalized learning pathways and supporting the creation of tangible, intellectual products, thereby "realizing Papert's vision" in a new technological era. Recent design-based studies have shown that building generative-AI tools functions as a constructionist learning activity in which students externalize domain knowledge through the design of AI systems (Huang et al., 2025a).

To understand the specific types of knowledge that must be integrated during this constructionist process, the Technological Pedagogical Content Knowledge (TPACK) framework is highly relevant. Building on the TPACK framework, recent research highlights the increasing need for educators and learners to integrate technological, pedagogical, and content knowledge specifically in AI-rich contexts, giving rise to expanded conceptions such as AI-TPACK (Kong et al., 2021; Ng et al., 2021). In this course, students are not just learning domain knowledge (CK); they are also learning how to teach that knowledge to an AI (a form of PK) using the affordances of a specific technology (GPTs, as TK). The student-built GPTs are the ultimate expression of this synthesis, representing the "TPACK" intersection where all three knowledge domains converge. The emergence of an "AI-TPACK" framework further underscores the need for a creative and strategic integration of AI technology with pedagogical and content goals.

Finally, Distributed Cognition theory provides a crucial lens for understanding the nature of the resulting human-AI partnership. Recent studies on human–AI collaboration conceptualize cognition as distributed across learners and intelligent systems, emphasizing how AI tools can function as cognitive partners that support, rather than replace, human judgment and expertise (Holstein et al., 2019; Celik et al., 2023). Recent studies on human–AI collaboration conceptualize cognition as distributed across learners and intelligent systems, emphasizing how AI tools can function as cognitive partners that support, rather than replace, human judgment and expertise (Holstein et al., 2019; Celik et al., 2023). The theory conceptualizes cognitive processes as inherently spread across individuals, artifacts, and environments, rather than being confined solely within the human mind. In this view, the custom GPT is not just a tool that a student uses; it becomes an integral part of the student's extended cognitive system. The student's expertise is distributed between their own mind and the logic they have programmed into the AI. For example, when the Interview Companion GPT identifies a missed follow-up question, it is the student's own research judgment, now offloaded and operationalized in the AI, that is performing the cognitive work. This theory explains why the AI feels like a "teammate"—it is a cognitive partner in a distributed system of intelligence.

Together, these four theories form a comprehensive foundation for the "Tool-to-Teammate" strategy. SDT explains the motivational engine, Constructionism explains the power of the making process, TPACK provides a map of the knowledge domains to be integrated, and Distributed Cognition explains the fundamental nature of the human-AI collaboration. This multi-theoretical perspective offers a rich and nuanced understanding of how students transition from passive users to active creators and co-cognitive partners with AI.

### Methodology: A Multi-Case Study Approach

To investigate the transformation of students from passive AI users to active creators, this research employed a qualitative multi-case study methodology. This approach is particularly suited for exploring complex phenomena within their real-life contexts and is ideal for generating rich, detailed insights into the processes and outcomes of innovative pedagogical strategies. The unit of analysis was the 'Creating the Frontier of No-code Smart Cities' course at the Singapore University of Technology and Design (SUTD), with three distinct student-built GPT-based projects serving as the focal cases for in-depth examination. These cases were selected for their diversity in application domain (qualitative research methods, urban observation, and pedagogical support) and their clear documentation of the transformation from an initial "tool" conception to a final "teammate" realization.

Data collection for this study was triangulated from multiple sources to ensure validity and depth. First, extensive documentation of the three case studies was gathered, including project reports, design specifications, and the final deployed GPT tools. This provided a detailed record of the students' design choices, AI implementations, and domain integrations. Second, interviews were conducted with the student project leads and instructors involved in the course. These semi-structured interviews probed the motivations behind specific design decisions, the challenges encountered during the development process, and the perceived learning outcomes for the students. Third, reflective essays written by the students as part of their course assessment were analyzed. These texts offered valuable first-person accounts of the students' thought processes, struggles, and epiphanies throughout the project lifecycle. Finally, contextual information regarding the course structure, the "Tool-to-Teammate" instructional strategy, and the theoretical underpinnings was synthesized from available course materials and relevant academic literature. The instructional design builds on earlier work proposing the "Tool-to-Teammate" strategy, in which students transition from passive users of generative AI to active designers of AI-powered learning tools (Sockalingam et al., 2025).

The analytical process followed a systematic procedure involving both within-case and cross-case analysis. Within each case, the data was coded to identify key themes related to the three core perspectives: Design, AI, and Domain Knowledge. For each perspective, the "Before" state (where AI was used as a generic tool) was contrasted with the "After" state (where AI was transformed into a customized teammate). This allowed for a granular deconstruction of the transformation process for each project. For example, for the Interview Companion GPT, the analysis would detail the limitations of standard text-based interactions ("Before") versus the implementation of a multimodal interface with live audio feedback ("After"), and how this design choice enabled the programming of persona-based logic to simulate challenging interviewees. This deep-dive approach ensured a comprehensive understanding of the unique dynamics of each case.

Following the individual case analyses, a cross-case synthesis was conducted to identify universal patterns and overarching themes. This synthesis focused on the interplay between the three analytical perspectives across all three cases. The goal was to move beyond isolated examples to construct a generalized model of how students leverage design, AI, and domain knowledge to foster their own agency. This process involved comparing and contrasting the findings from each case to highlight recurring mechanisms, such as how mastery of domain knowledge invariably informs the AI's architecture, or how specific design choices serve to scaffold human learning rather than automate it. By combining deep-dive analysis with cross-case synthesis, this methodology provides a holistic and nuanced understanding of the pedagogical transformation at SUTD, offering both rich descriptive detail and generalizable theoretical insights for the field of AI in education.

## Findings

This section presents a detailed analysis of the three case studies from the 'No-code Smart Cities' course, examining the transformation of students from passive AI users to active creators. The analysis is structured around the dichotomy of the "Before" state, where AI functioned as a generic tool, and the "After" state, where students engineered their own AI as a specialized teammate. Each case demonstrates a unique application of the interplay between Design, AI, and Domain Knowledge.

## Case Study 1: The Interview Companion GPT

The Interview Companion was developed to address the difficulty of practicing high-stakes qualitative research interviews. It serves as a multimodal simulator that offers feedback on transcripts and live audio sessions.

**Comparison: The Practice of Research Interviewing**

| Perspective | Pre-AI Era (Manual Learning) | Custom GPT Era (AI as Teammate) |
|---|---|---|
| Design | **Static & Peer-Dependent:** Practice relied on role-playing with peers or professors, which was time-bound and often lacked rigorous, objective critique. | **Interaction Design:** A multimodal system designed for "simulator-grade" rehearsal, providing immediate feedback on pacing, tone, and question framing. |

| AI | Non-existent: Technology was used only for recording (audio/video), requiring manual transcription and human analysis. | Agentic Logic: Uses sophisticated "persona-based" prompting to simulate challenging interviewees and identify missed follow-up opportunities. |
|---|---|---|
| Domain | Linear Theory Application: Students read about interviewing techniques but could only apply them in high-stakes, real-world scenarios. | Constructionist Mastery: Students "encode" interview frameworks into the GPT, mastering domain expertise by becoming the "teacher" of the AI. |

**Detailed Analysis**

- **Design:** The transition represents a move from "learning about" to "designing for." Students identified that generic AI lacked "criticality," leading them to design an interface that toggles between formal and conversational modes.

- **AI:** The AI functions as a "glass box." By configuring the system instructions, students learned to manage AI drift and established functional boundaries to ensure the AI remains in its assigned role (e.g., mock interviewee vs. feedback provider).

- **Domain Knowledge:** This tool transforms qualitative research from an abstract theory into a repeatable skill. By teaching the AI to probe for deeper insights, students reinforce their own understanding of research intent.

**Real-World Deployment and Educational Impact**

The Interview Companion GPT was not only tested as a classroom prototype but was also deployed in authentic research practice. Students from the Master of Science in Urban Science, Policy and Planning (MUSPP) program used the tool during a collaboration with the Oral History Centre at the National Archives of Singapore (SUTD Urban Science, 2024). In this context, the GPT served as a real-time cognitive companion that supported students in conducting oral history interviews by helping them reflect on questioning strategies, identify missed follow-ups, and maintain conversational flow.

This real-world deployment demonstrates how student-designed AI tools can extend beyond classroom experimentation into professional practice. Rather than replacing the interviewer, the GPT functioned as a reflective partner that augmented the interviewer's awareness and decision-making during the interview process. The experience illustrates the principles of Human-Centered AI, where AI is designed to enhance human judgment rather than automate complex human skills.

From a pedagogical perspective, the collaboration with the National Archives provided students with an authentic research environment in which their custom GPT tool could be tested and refined. This experience strengthened students' understanding of qualitative interviewing as both a technical skill and a reflective practice. It also reinforced the constructionist learning process described in this study: students deepen their domain expertise when they design AI systems that operationalize that expertise in real-world contexts.

The deployment of the Interview Companion GPT in oral history interviews therefore provides an important example of how student-created AI tools can bridge the gap between academic learning and professional practice, demonstrating the broader educational value of the "Tool-to-Teammate" approach.

## Case Study 2: The Urban Observer GPT

This case study examines how students moved beyond "passive looking" to a "structured seeing" model, using AI to augment field observation in urban design.

### Comparison: The Field Observation Process

| Perspective | Pre-AI Era (Manual Learning) | Custom GPT Era (AI as Teammate) |
|---|---|---|
| Design | **Manual & Disorganized:** Observations were captured in raw field notes, often leading to information overload and the omission of critical sensory data. | **Guided Pedagogy:** Designed as a "Friendly Tutor" that uses a Socratic questioning method to guide students through a structured observation journey. |
| AI | **Passive Documentation:** Technology served only as a camera or notepad with no analytical or guiding capability in the field. | **Vision-to-Framework:** Utilizes computer vision and structured data processing to help students categorize spatial and social data into an "Observation Matrix." |
| Domain | **Framework Gap:** Students often struggled to apply urban theories (e.g., Creswell or Gehl) to the chaotic, real-time environment of a city. | **Theoretical Encoding:** The GPT forces students to apply theoretical layers to their observations, bridging the gap between classroom theory and field practice. |

### Detailed Analysis

- **Design:** The "Friendly Tutor" archetype is a specific design choice that prevents the AI from giving away answers, ensuring the student remains the primary cognitive actor.

- **AI:** The project highlights the management of technical frontiers, such as using AI to generate "future-state imagery" from static photos, allowing students to visualize urban interventions.

- **Domain Knowledge:** Urban designers must learn to see "layers" (physical vs. social). The tool aids this by acting as a cognitive partner that tracks environmental details while the student focuses on social behavior.

**Case Study 3: Buddy Buddy (Flipped Classroom)**
Buddy Buddy addresses the "Flipped Classroom" challenge by personalizing pre-class study and providing instructors with insights into student preparation.

**Comparison: Pre-Class Preparation and Engagement**

| Perspective | Pre-AI Era (Manual Learning) | Custom GPT Era (AI as Teammate) |
|---|---|---|
| Design | **Information Intensive:** Pre-class work involved reading static PDFs. There was zero visibility for the teacher regarding student comprehension. | **Interaction Intensive:** Designed to gather student backgrounds and CVs, highlighting how individual experiences contribute to the upcoming lesson. |
| AI | **Non-existent:** Educational technology was limited to static Learning Management Systems (LMS) with no conversational or adaptive capacity. | **Personalization Engine:** Acts as an "upskilling-as-a-service" partner, creating conceptual bridges between a student's past work and new course content. |
| Domain | **Disconnected Theory:** Students with professional experience often felt academic content was redundant or disconnected from their real-world skills. | **Knowledge-Building:** The GPT helps students translate their personal expertise into the domain frameworks of the course, fostering "Social Metacognition." |

**Detailed Analysis**
- **Design:** The tool creates a "closed-loop" system where the student's private interaction is summarized into an "Educator Report," informing the teacher's classroom strategy.

- **AI:** Buddy Buddy moves AI beyond Retrieval-Augmented Generation (RAG) into "Interaction-Intensive" territory, where the value is in the dialogue rather than just the retrieved fact.

- **Domain Knowledge:** The domain of "Smart Cities" is inherently multidisciplinary. Buddy Buddy helps students from various backgrounds (architecture, engineering, policy) find their unique "entry point" into the curriculum.

**Cross-Case Synthesis: The Synergy of Design, AI, and Domain Knowledge**

A cross-case synthesis of the three projects—the Interview Companion GPT, the Urban Observer GPT, and Buddy Buddy—reveals a powerful and consistent learning model driven by the dynamic interplay between Design, Artificial Intelligence, and Domain Knowledge. This model, which can be conceptualized as a "Trilingual triad," demonstrates how these three elements are not independent variables but deeply interconnected components of a unified system of learning. The transformation from AI as a passive tool to AI as an active teammate is not the result of mastering one of these areas in isolation, but rather the outcome of orchestrating their synergy. This synthesis uncovers universal patterns that illuminate the pedagogical power of student-led AI creation projects.

First, a foundational principle emerges: Domain Knowledge informs AI Architecture. Across all three cases, the successful creation of a useful AI teammate began with a deep and explicit understanding of the domain's rules, principles, and analytical frameworks. To build the Interview Companion, students had to codify the unwritten rules of skilled interviewing, such as recognizing when a follow-up question is needed. To create the Urban Observer, they had to translate abstract urban planning theories into a structured data-categorization matrix. And for Buddy Buddy, they had to understand the course's conceptual frameworks well enough to help students map their personal experiences onto them. This process forces a profound level of metacognition, where students must externalize their implicit, tacit knowledge into explicit, teachable criteria that can be "programmed" into the AI. The AI, therefore, becomes a living artifact of the student's own mastery of the domain.

Second, Design Bridges Theory and Practice. The technical capabilities of AI models are vast, but their educational utility depends entirely on how they are mediated by thoughtful design. The students in these case studies acted as skilled designers, crafting interfaces, workflows, and interaction patterns that channeled the AI's power toward specific pedagogical goals. The multimodal interface of the Interview Companion was not just a feature; it was a design choice to create a more authentic simulation of reality. The "Friendly Tutor" persona of the Urban Observer was a strategic decision to manage cognitive load and scaffold the learning process. The interactive onboarding flow of Buddy Buddy was designed to shift the focus from information delivery to value recognition. These design choices are what differentiate a mere tool from a true teammate. Good design ensures that the AI reinforces, rather than replaces, human learning and critical thinking.

Third, AI Empowers Agency. The ultimate outcome of this synergistic process is the empowerment of student agency. This shift toward student agency aligns with recent studies in engineering and design education demonstrating that generative AI can reposition learners as co-creators rather than passive consumers of technology (Willems et al., 2025).By transitioning from being users who must adapt to the AI, to creators who shape the AI to fit their needs, students gain a profound sense of competence and autonomy. This act of creation is a powerful affirmation of their ability to influence and control their learning environment. They are no longer passive consumers of technology but its conscious and critical architects. This aligns perfectly with the predictions of Self-Determination Theory, where the satisfaction of the psychological needs for autonomy and competence is a primary driver of intrinsic motivation and engagement. The creation of these tools provides tangible proof of their capabilities, boosting their confidence and fostering a more proactive and self-directed approach to learning. This progression from consumption to curation and finally to creation reflects a trajectory of increasing digital and AI literacy, moving students toward the status of "AI literate designers" who can ethically and effectively wield these powerful tools.

In essence, the synthesis reveals a virtuous cycle. Mastery of the domain provides the substance (what the AI knows). Thoughtful design provides the form (how the AI interacts). And the act of building this symbiotic system provides the catalyst for student empowerment (agency). This triad model offers a robust conceptual framework for designing future curricula that integrate no-code AI development, demonstrating that the most effective learning occurs not when students simply use AI, but when they learn to collaborate with it as a teammate.

**Conclusion**

In conclusion, the 'No-code Smart Cities' course at SUTD offers a vital glimpse into a future of education where students are empowered as creators and critical partners in the age of AI. By embracing a pedagogy grounded in the synergy of domain knowledge, thoughtful design, and AI capabilities, and guided by a robust multi-theoretical foundation, educators can foster a generation of learners who are not just prepared to use AI, but to shape it responsibly and effectively.

Through detailed deep dives into three distinct case studies—the Interview Companion GPT, the Urban Observer GPT, and Buddy Buddy—the study demonstrated that the act of creation is the central engine of this transformation. The analysis revealed a powerful synergistic model, or "Trilingual triad," (Figure 1) wherein mastery of domain knowledge informs the AI's architecture, thoughtful design mediates the AI's capabilities for educational benefit, and the resulting tool empowers students with a heightened sense of agency and competence. Students were not merely learning about AI; they were learning *by* applying their domain expertise to architect novel solutions to authentic problems, thereby achieving a profound and durable understanding of their subject matter.

The cross-case synthesis indicates that this model offers a replicable and robust framework for integrating no-code AI development into higher education. It provides a practical pathway for cultivating a more sophisticated form of AI literacy—one that extends beyond operational proficiency to encompass critical evaluation, ethical consideration, and creative application. By positioning students as co-designers and collaborators with AI, this pedagogical approach moves beyond simplistic narratives of automation and disruption, instead fostering a partnership between human ingenuity and artificial intelligence. The findings suggest that future curricula should prioritize such hands-on, project-based experiences to prepare learners for a world where the ability to create and critically engage with AI is not a niche skill but a fundamental competency.

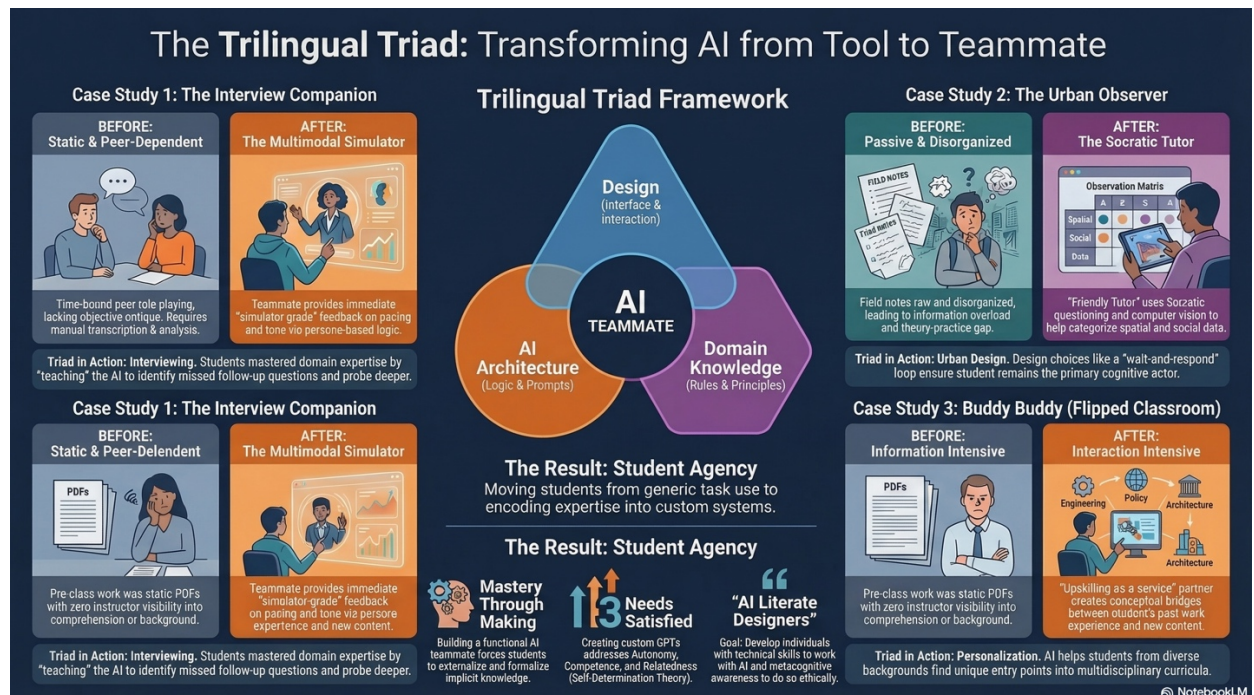

Figure 1. The "Trilingual Triad" generated by NotebookLM

## Design Principles

### An Integrated Framework of Human-Centered AI Principles
The student-led creation of GPT-based teammates in the 'No-code Smart Cities' course operationalizes key human-centered AI principles by weaving them into the fabric of Design, AI, and Domain Knowledge. This creates a self-reinforcing system where ethical and effective AI use is an emergent property of the design process itself.

### Augmenting, Not Replacing, Human Capabilities
This foundational principle is achieved through a deliberate alignment of all three pillars. Domain Knowledge defines the core human skill that must be preserved and enhanced— for instance, the nuanced art of qualitative interviewing or the critical practice of urban observation. The AI architecture is then explicitly designed to support this skill, not to perform it on the user's behalf. For example, the AI is programmed to identify a *missed* follow-up question, not to generate the perfect one. Finally, the Design of the interface enforces this boundary; the Urban Observer's "wait-and-respond" loop is a design feature that physically prevents the AI from jumping ahead, thereby forcing the user to engage in the primary cognitive work. In this way, augmentation is not a policy but a product of the integrated system.

### Ensuring Human Oversight and Control

Human control is embedded directly into the creative process. Students begin with their Domain Knowledge, which provides the authoritative standard against which the AI's behavior is judged. They then exercise control by using AI techniques like constraint-based prompting and persona engineering to encode this domain expertise into the model's logic. This act of "teaching the AI" is the ultimate form of oversight. The Design of the tool further reinforces this control by making the AI's functions transparent and modifiable. A student can see the prompt that defines the mock interviewee's persona and adjust it if the simulation becomes unrealistic. Thus, control is not a post-hoc check but a continuous, integral part of the co-design workflow.

### Empowering Learners as Co-Creators

Empowerment is the direct result of the learner's journey across the three domains. To build a teammate, a student must first master their Domain Knowledge deeply enough to formalize it. They then acquire AI literacy by learning to translate this knowledge into a functional system. Finally, they develop Design thinking to create an interface that effectively mediates the interaction between the user and the AI. This holistic process transforms the learner from a passive recipient of technology into an active architect, granting them the competence and autonomy that are the hallmarks of true empowerment. The final product—a custom GPT—is a tangible artifact of their agency.

### Fostering Flourishing, Relationships, and Creativity

These higher-order human goals are the natural outcomes of the synergistic triad. The deep engagement required to integrate Domain Knowledge with AI and Design is an inherently creative act, fostering intellectual flourishing. The collaborative nature of the projects builds relationships among students who share in the complex work of co-creation. Furthermore, the resulting tools often serve to strengthen the relationship between the student and their discipline, as seen in Buddy Buddy, which helps students connect their personal experiences to theoretical frameworks. In this model, creativity, community, and a deeper connection to one's field of study are not accidental byproducts but central features of the learning experience.

In summary, this framework shows that the principles of human-centered AI are not external constraints to be applied after a system is built. Instead, they are the very essence of a well-designed pedagogical process that intentionally integrates deep domain expertise, thoughtful AI architecture, and purposeful design. The "Tool-to-Teammate" strategy at SUTD provides a powerful blueprint for how this integration can be achieved in practice.

### Declaration

This study was conducted through human-AI collaboration. While the authors developed the original research concepts of research (including all the case studies), lesson plans for the "Creating the Frontier of No-code Smart City" course, several AI models, including

ChatGPT, Gemini, NotebookLM, Qwen, and Kimi, were utilized to assist with technical writing and diagram design. The core intellectual property and course delivery remain the sole work of the authors.

## References

Bialik, M., & Fadel, C. (2018). *Skills for the AI age: What students should learn*. Center for Curriculum Redesign.

Celik, I., Mertens, A., & Huang, J. (2023). Human–AI collaboration in education: A systematic review of empirical studies. *Computers & Education: Artificial Intelligence, 4*, 100115. https://doi.org/10.1016/j.caeai.2023.100115

Holmes, W., Bialik, M., & Fadel, C. (2019). *Artificial intelligence in education: Promises and implications for teaching and learning*. Center for Curriculum Redesign.

Holstein, K., McLaren, B. M., & Aleven, V. (2019).
Designing for complementarity: Teacher and AI roles in orchestrating classroom learning. In *Proceedings of the 2019 CHI Conference on Human Factors in Computing Systems* (pp. 1–13). https://doi.org/10.1145/3290605.3300762

Huang, Q., & Poon, K. W. (2025). Human and AI collaboration in fitness education: A longitudinal study with a Pilates instructor. *arXiv preprint*. https://arxiv.org/abs/2506.06383

Huang, Q., & Willems, T. (2025). Empowering educators in the age of AI: An empirical study on creating custom GPTs in qualitative research method education. *arXiv preprint*. https://arxiv.org/abs/2507.21074

Huang, Q., Willems, T., & Poon, K. W. (2025). The application of GPT-4 in grading design university students' assignments: An exploratory study. *International Journal of Technology and Design Education*.

Huang, Q., Sockalingam, N., Willems, T., & Poon, K. W. (2025a). Designing knowledge tools: How students transition from using to creating generative AI in STEAM classrooms. *arXiv preprint*. https://doi.org/10.48550/arXiv.2510.19405

Kasneci, E., Sessler, K., Küchemann, S., Bannert, M., Dementieva, D., Fischer, F., … Kasneci, G. (2023). ChatGPT for good? On opportunities and challenges of large language models for education. *Learning and Individual Differences, 103*, 102274. https://doi.org/10.1016/j.lindif.2023.102274

Kong, S. C., Cheung, W. M. Y., & Zhang, G. (2021). Development of computational thinking and AI literacy: A systematic review. *Educational Technology & Society, 24*(3), 1–15.

Long, D., & Magerko, B. (2020). What is AI literacy? Competencies and design considerations. In *Proceedings of the 2020 CHI Conference on Human Factors in Computing Systems* (pp. 1–16). https://doi.org/10.1145/3313831.3376727

Ng, D. T. K., Leung, J. K. L., Chu, S. K. W., & Qiao, M. S. (2021). Conceptualizing AI literacy: An exploratory review. *Computers and Education: Artificial Intelligence, 2*, 100041. https://doi.org/10.1016/j.caeai.2021.100041

Shneiderman, B. (2020). Human-centered artificial intelligence: Reliable, safe & trustworthy. *International Journal of Human–Computer Interaction, 36*(6), 495–504. https://doi.org/10.1080/10447318.2020.1741118

Sockalingam, N., Huang, Q., Willems, T., & Poon, K. W. (2025). Innovative "tool-to-teammate" instructional strategy for using ChatGPT in higher education. In *EDULEARN25 Proceedings* (pp. 7891–7900).

SUTD Urban Science. (2024). *MUSPP students shine at the Oral History Centre National Archives with CustomGPT Interview Companion.* https://urbanscience.sutd.edu.sg/news-and-events/muspp-24-students-shine-at-the-oral-history-centre-national-archives-with-customgpt-interview-companion/

Willems, T., Khan, S., Huang, Q., Camburn, B., Sockalingam, N., & Poon, K. W. (2025). To use or to refuse? Re-centering student agency with generative AI in engineering design education. In *IEEE International Conference on Teaching, Assessment, and Learning for Engineering*.

Zawacki-Richter, O., Marín, V. I., Bond, M., & Gouverneur, F. (2019). Systematic review of research on artificial intelligence applications in higher education. *International Journal of Educational Technology in Higher Education, 16*(1), 39. https://doi.org/10.1186/s41239-019-0171-0